\documentclass{article}

    \usepackage[preprint, nonatbib]{neurips_2022}

\usepackage[utf8]{inputenc} 
\usepackage[T1]{fontenc}    
\usepackage{hyperref}       
\usepackage{url}            
\usepackage{booktabs}       
\usepackage{amsfonts}       
\usepackage{nicefrac}       
\usepackage{microtype}      
\usepackage{xcolor}         

\usepackage{amsmath}
\usepackage{amssymb}
\usepackage{mathtools}
\usepackage{amsthm}
\usepackage{caption}
\usepackage{subcaption}

\newcommand{\el}{\textit{~et~al.}}

\title{A Computational Framework of Cortical \\
Microcircuits Approximates Sign-concordant \\
Random Backpropagation}

\author{%
  Yukun Yang\thanks{Y. Yang is an incoming Ph.D. student to University of California, Santa Barbara.} \\
  ECE Department\\
  University of California, Santa Barbara\\
  Santa Barbara, CA 93106 \\
  \texttt{yukunyang@ucsb.edu} \\
  \And
  Peng Li \\
  ECE Department\\
  University of California, Santa Barbara\\
  Santa Barbara, CA 93106 \\
  \texttt{lip@ucsb.edu} \\
}

\begin{document}

\maketitle

\begin{abstract}
Several recent studies attempt to address the biological implausibility of the well-known backpropagation (BP) method. While promising methods such as feedback alignment, direct feedback alignment, and their variants like sign-concordant feedback alignment tackle BP's weight transport problem, their validity remains controversial owing to a set of other unsolved issues. In this work, we answer the question of whether it is possible to realize random backpropagation solely based on mechanisms observed in neuroscience. We propose a hypothetical framework consisting of a new microcircuit architecture and its supporting Hebbian learning rules. Comprising three types of cells and two types of synaptic connectivity, the proposed microcircuit architecture computes and propagates error signals through local feedback connections and supports the training of multi-layered spiking neural networks with a globally defined spiking error function.  We employ the Hebbian rule operating in local compartments to update synaptic weights and achieve supervised learning in a biologically plausible manner. Finally, we interpret the proposed framework from an optimization point of view and show its equivalence to sign-concordant feedback alignment. The proposed framework is benchmarked on several datasets including MNIST and CIFAR10,  demonstrating promising BP-comparable accuracy.
\end{abstract}

\section{Introduction}
\label{sec:intro}
Artificial intelligence has gained remarkable achievements in recent years \cite{Hinton06, Bengio+chapter2007, schmidhuber2015deep, goodfellow2016deep}, and learning by backpropagation (BP) \cite{rumelhart1986learning} is one of the most popular methods. However, BP is generally believed  to be impossible in the brain for many reasons \cite{stork1989backpropagation,  chinta2012adaptive, lillicrap2016random, nokland2016direct, illing2019biologically}, among these, one key reason is that it requires symmetrical weights between forward and backward paths (also known as the weight transport problem) to propagate correct error information \cite{stork1989backpropagation, chinta2012adaptive, grossberg1987competitive, ororbia2017learning}. 
Many works tried to mitigate this constraint, including feedback alignment (FA) \cite{lillicrap2016random} and direct feedback alignment (DFA) \cite{nokland2016direct}. Both FA and DFA used random weights to backpropagate error information, where FA and its variants \cite{liao2016important, moskovitz2018feedback, akrout2019deep} backpropagate error layer-by-layer, and DFA broadcasts error to each layer directly.

Both FA and DFA interestingly revealed and brought much attention to how brains might implement BP. However, there are still many unsolved issues towards fully bio-plausible algorithms. We conclude these issues as the followings:

\textbf{1)} BP uses different rules in forward- and backward-propagation. The forward signal goes through a non-linear function, while the backward signal is multiplied by the derivative of this non-linear function \cite{sacramento2018dendritic, payeur2021burst}. 
\textbf{2)} BP requires activation and error signals to coexist in one neuron. New insights regarding how neuron achieves this remain as an important yet underdeveloped topic \cite{sacramento2018dendritic, kording2001supervised, spratling2002cortical, spratling2006feedback, spruston2008pyramidal,guerguiev2017towards, neftci2017event}.
\textbf{3)} BP asks for separate forward and backward phases. A bio-plausible approach should eliminate this constraint like previous works \cite{sacramento2018dendritic, bellec2020solution}.
\textbf{4)} BP usually violates Dale's principle \cite{dale1935pharmacology, eccles1954cholinergic}. Dale's principle requires all synapses from the same cell to have the same sign - a strong biological constraint that can compromise network performance \cite{moskovitz2018feedback} and has been omitted in previous works \cite{lillicrap2016random, moskovitz2018feedback, sacramento2018dendritic, payeur2021burst, zenke2018superspike}. Cornford\el \cite{cornford2021learning} tackled this issue on artificial neural networks (ANNs). 
\textbf{5)} BP does not take into account the existence of various neuron types \cite{bartunov2018assessing}. Understanding how different types of biological neurons cooperate to enable learning is fundamentally important in neuroscience \cite{sacramento2018dendritic, payeur2021burst}.
\textbf{6)} BP lacks spike-timing level understanding \cite{payeur2021burst, brette2015philosophy, yang2021backpropagated}. 

Instead of revising BP, we integrate biological observations and devise a hypothetical framework consisting of a new architecture and its supporting learning rules that can address all of these major issues above. We demonstrate that it is possible to realize bio-plausible random error backpropagation entirely based on neuroscience facts. we summarize our main contributions as follows:


\begin{itemize}
\item We hypothesize several microcircuit architectures including various types of spike-timing based neurons and their special connectivity abiding by Dale's Principle.
\item We enable error backpropagation with spikes and provide a new model of how neurons achieve this through their internal dynamics.
\item We present a pointwise SOM model, and a supporting assembly competition mechanism to form pairs between SOM and Pyr cells automatically.
\item We propose a general local learning rule, which applies \textbf{Hebbian} and \textbf{anti-Hebbian} learning on synapses that connect to \textbf{basal} and \textbf{apical} dendrites respectively.
\end{itemize}

\section{Background}
\label{sec:background}

\subsection{The spiking neuron model}
\label{sec:background_snn}

All cell types are modeled by the leaky integrate-and-fire (LIF) neuron model \cite{gerstner2002spiking}, one of the most common choices for describing the dynamics of spiking neurons. Importantly, LIF neurons are able to fire spikes with accurate timing, which is one of the major differences between our work and previous works using rate-based neurons \cite{sacramento2018dendritic, bastos2012canonical, whittington2017approximation}.

The dynamics of somatic membrane potential $u$ of neuron $i$ in layer $l$ is described by:
\begin{equation}
    \tau_m\frac{d {u}_i^{(l)}(t)}{d t}  = -{u}_i^{(l)}(t)+ I_i^{(l)}(t)
    + \eta_i^{(l)}(t),
    \label{eq:pyramidal}
\end{equation}
where $I_i^{(l)}$ is the total input of synaptic currents onto neuron's soma, and $\eta_i^{(l)}(t)$ denotes the reset function. A spiking neuron reset its membrane potential to the resting potential $v_{\rm rest}$ (we set $v_{\rm rest}=0$) each time when it fires a spike. We model $\eta_i^{(l)}(t)$ as the time convolution (*) between a reset kernel $\nu$ and the neuron's output spike train: $\eta_i^{(l)}(t) = (\nu \ast {s}_i^{(l)}) (t)$. The reset kernel $\nu(t)=-\vartheta\delta(t)$. The amount of resetting is equal to the threshold $\vartheta$ (we set $\vartheta=1$), and $\delta(t)$ is the Dirac delta function. The neuron's output spike train is also modeled by a serious of delta functions as: ${s}_i^{(l)}=\sum_f\delta(t-t_{i(f)}^{(l)}).$

Here, $t_{i(f)}^{(l)}$ represent the firing time of the $f^{th}$ spike of neuron $i$ in layer $(l)$. An output spike is generated once the membrane potential reaches the threshold $\vartheta$.

We model the general total input current on neuron i as:$I_i = \sum_{j}{w_{ij}} a_j$. where the postsynaptic current (PSC) $a_j$ generated from neuron $j$ can be expressed by:
\begin{equation}
    a_j(t)= \left\{
    \begin{array}{lr}
    (s_j*\epsilon)(t),& {\rm Excitatory}\\
    {-(s_j*\epsilon)(t)},& {\rm Inhibitory}
    \end{array}\right.
\label{eq:psc_conv}
\end{equation}
where (*) represents the time convolution. $\epsilon(t)= ({1}/{\tau_s})\cdot exp{(-t/{\tau_s})} H(t)$ is the impulse response. $H(\cdot)$ represents the Heaviside step function: $H(t)=1, t\geq 0$ and  $H(t)=0, t<0$. Following Dale's principle, we force all synaptic weights to be positive $({w_{ij}}>0)$, and separate excitatory and inhibitory neurons by the sign of PSCs.

\subsection{Backpropagation of spiking neural networks}
\label{sec:bp_snn}
Recent works demonstrate the great potential of BP in training Spiking Neural Networks (SNNs) \cite{zenke2018superspike, yang2021backpropagated, shrestha2018slayer, zhang2020temporal}. 

We conclude a general approach as following:

BP in SNNs involves error propagation in two directions: along the time axis (also known as the temporal credit assignment problem), and alone the layers. According to (\ref{eq:psc_conv}), a neuron's PSC is modeled by the unit impulse response of its output spike train with kernel $\epsilon$. Therefore convolving $\partial L/\partial a(t)$ with a time-reversed kernel $\Tilde{\epsilon}(t) = \epsilon(-t)$ distribute the error to all spike trains fired before the current time t \cite{shrestha2018slayer}. Further, BP-based methods usually introduce a surrogate term to handle the spiking neurons' non-differentiable firing relationship between membrane potentials and spike trains, which approximate $\partial L/\partial u(t)$ as $\sigma'(u(t))\cdot\partial L/\partial s(t)$ \cite{zenke2018superspike}.

Concatenating all signals of the same layer as a vector, we denote $\partial L/\partial \boldsymbol{u}^{(l+1)}(t)$ as $\boldsymbol{\delta}_i^{(l+1)}(t)$. BP propagates it to previous layers by:$\partial L/\partial \boldsymbol{a}^{(l)}(t) = (\boldsymbol{\rm W}^{(l+1)})^T\boldsymbol{\delta}_i^{(l+1)}(t)$, which is similar to how BP works in ANNs.

\subsection{Sign-concordant feedback alignment}

Sign-concordant feedback alignment (SFA) algorithm \cite{liao2016important, moskovitz2018feedback}, which requires symmetric forward and backward weight signs, targets to solve BP's bio-implausible weight transport problem. SFA outperforms FA, especially on deep convolutional neural networks. However, its assumptions about symmetric signs are controversial. We discuss both the weight transport problem of BP and SFA here: 
Using $\boldsymbol{x}^{(l+1)}$ and $\boldsymbol{y}^{(l+1)}$ to denote the total input and output vectors of layer $(l+1)$ in a non-spiking artificial neural network (ANN), then $\boldsymbol{y}^{(l+1)} = \sigma(\boldsymbol{x}^{(l+1)})$, where $\sigma$ is a non-linear function \cite{rumelhart1986learning, nair2010rectified}. The total input $\boldsymbol{x}^{(l+1)}=\boldsymbol{\rm{W}}^{(l+1)}\boldsymbol{y}^{(l)}+\boldsymbol{b}^{(l+1)}$, where $\boldsymbol{\rm{W}}^{(l+1)}$, $\boldsymbol{b}^{(l+1)}$ are the forward weights and bias in layer $(l+1)$ respectively. For a given loss function $L$, define error signals $\boldsymbol{\delta}^{(l+1)}=\partial L/\partial \boldsymbol{x}^{(l+1)}$. BP propagates $\boldsymbol{\delta}^{(l+1)}$ to $\boldsymbol{y}^{(l)}$ by: 
\begin{equation}
    \partial L/ \partial \boldsymbol{y}^{(l)} = (\boldsymbol{\rm{W}}^{(l+1)})^T \boldsymbol{\delta}^{(l+1)}
\end{equation}
The backward weight $(\boldsymbol{\rm{W}}^{(l+1)})^T$ is the transpose of the forward weights. However, biology synapses propagate information in a single direction, so the weight transport requires the corresponding forward and backward synapses to have the same value, which is not bio-plausible.
FA uses a random matrix $\boldsymbol{\rm{B}}$ to substitute $\boldsymbol{\rm{W}}^T$ (Omitting the layer index for short). As stated in Lillicrap\el~\cite{lillicrap2016random}, any matrix $\boldsymbol{\rm{B}}$ that on average suffice $\boldsymbol{\delta}^T\boldsymbol{\rm{W}}\boldsymbol{\rm{B}}\boldsymbol{\delta}>0$ supports successful training. Geometrically, if the signal $\boldsymbol{\rm{B}}\boldsymbol{\delta}$ lies within 90\textdegree of the correct derivative $\boldsymbol{\rm{W}}^T\boldsymbol{\delta}$ on average, it can propagate useful error information backwards.

However, FA is notorious for its poor performance on complex tasks, especially when applied to training deep convolutional neural networks \cite{liao2016important, moskovitz2018feedback, akrout2019deep}. To mitigate this issue, SFA was proposed \cite{liao2016important, moskovitz2018feedback}, which requires ${\rm sign}(\boldsymbol{\rm{B}}) = {\rm sign}(\boldsymbol{\rm{W}}^T)$. Because each corresponding elements of $\boldsymbol{\rm{B}}$ and $\boldsymbol{\rm{W}}^T$ have the same sign, SFA always guarantees $\boldsymbol{\delta}^T\boldsymbol{\rm{W}}\boldsymbol{\rm{B}}\boldsymbol{\delta}>0$.

\section{Microcircuits}

\begin{figure}[t]
    \centering

    \includegraphics[width=0.95\textwidth]{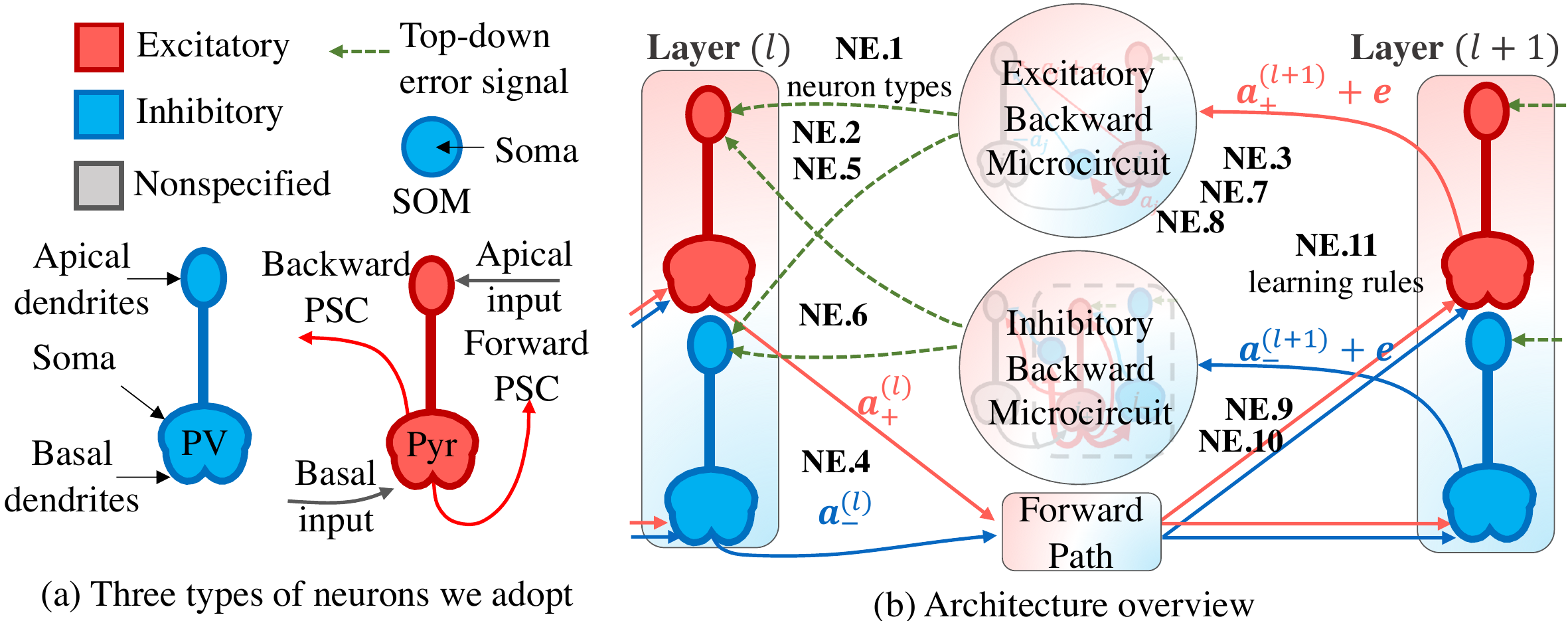}
    \caption{Three types of neurons' model and the architecture between two layers (We put the 12 neuroscience experimental observations near the components that reference them).}
    \label{fig:architecture}

    \vspace{-0.3cm}
\end{figure}

To achieve bio-plausible error processing, we identified several computational issues (\textbf{IS}) to solve (corresponding to the 6 bio-implausibilities of BP described in Section \ref{sec:intro}): 
\textbf{IS.1}: Apply unified neural dynamics rules to forward and backward input currents. 
\textbf{IS.2}: Store forward activation and backward error signals separately. Process error through interactions between different compartments.
\textbf{IS.3}: During continuous training, all types of synaptic plasticity are always enabled, with no separate forward-backward phases.
\textbf{IS.4}: Follow Dale's principle. A neuron is only allowed to send excitatory or inhibitory synapses.
\textbf{IS.5}: Realize forward and backward propagation with special cooperation between different types of neurons.
\textbf{IS.6}: Provide insights of the temporal credit assignment problem to neurons that can fire spikes with accurate timing. Achieve continuous valued error propagation through discontinuous all-or-none spikes.

Motivated by the recent developments in neuronscience (\textbf{NE}), we synthesize biological experimental evidence under a systematic computational framework to realize random error backpropagation. In this section, we introduce our microcircuit architecture and discuss its bio-plausibility.  

Guided by Dale's principle (\textbf{IS.4}) and based on observation \textbf{NE.1}: Neocortical visual processing neurons include excitatory pyramidal (Pyr) neurons and various types of inhibitory interneurons. The two most abundant types are parvalbumin-expressing (PV) neurons and somatostatin-expressing (SOM) neurons \cite{cottam2013target}, we build our microcircuits with three types of neurons: Pyr, PV, and SOM neurons (\textbf{IS.5}).

Undoubtedly, biology neurons' anatomies are complex. We follow the simplified multi-compartments Pyr model \cite{sacramento2018dendritic, payeur2021burst, spratling2002cortical, guerguiev2017towards}, and model PV cells in the same way. \textbf{NE.2}: Wouterlood\el \cite{wouterlood1995parvalbumin} states that PV neurons may have two types: the one with long dendrites and the one with short dendrites. Zipp\el \cite{zipp1989entorhinal} and Chiovini\el \cite{chiovini2014dendritic} provide observation of PV neurons with long apical dendrites. Our PV model follows this type. Yet \textbf{NE.3}: SOM cells usually have spatially constrained dendrites \cite{urban2016somatostatin}, so we model SOM cells by a single compartment. 

Regarding computational requirements, we divide the architecture into three main parts, as shown in Figure \ref{fig:architecture}: the forward path, the excitatory backward path, and the inhibitory backward path. The forward path shows the connections that propagate the activation of excitatory and inhibitory neurons to the next layer, and the two backward paths show the connections that propagate the errors of excitatory and inhibitory neurons backward, respectively. 

Since  microcircuit design is largely dependent on intraneural dynamics, we begin our discussion with modeling of the apical dendrites of spiking neurons. Then, we exhibit the forward, excitatory backward, and inhibitory backward paths one by one.

\subsection{Apical dendrites modeling}

Here we present a model that transmits a continuous error signal through discontinuous spikes (\textbf{IS.6}). To achieve this, we thought it might be sensible to distinguish between small and large signal characteristics, as is commonly done in analog circuit analysis, when modeling the relationship between apical dendritic currents and output PSCs of neurons. If the stimulation from apical dendrites is small, we would expect a linear addition assumption, where the slope depends on the current membrane potential. The closer the membrane potential is to the threshold, the more likely it is that additional stimulation will cause PSC changes, and thus the two signals, apical stimulation and PSC, are more correlated. Meanwhile, significant non-linearity would occur only if the increment is large (\textbf{IS.1}, \textbf{IS.2}). To provide more intuition, we begin this section with a preliminary single-neuron experiment.
\subsubsection{single neuron experiment}

Simulate a single LIF Pyr neuron. 50 randomly set excitatory/inhibitory neurons synapse to the neuron's basal dendrites and other 50 synapse to its apical dendrites. Firing behavior and synaptic weights are randomly assigned. We model the total input $I(t) =I_b(t)+I_a(t)$ as the sum of currents from basal dendrites $I_b(t)$ and apical dendrites $I_a(t)$. We first simulate the response of LIF neurons to $I_b(a)$, and name the resulting PSC $a_b(t)$. We then re-simulate on the exact same random seed with both inputs applied and name the resulting PSC $a(t)$.

We sample pairs of $I_a(t)$ and apical-induced PSC change $a(t) - a_b(t)$ at multiple time points, and plot in Figure \ref{fig:apical} (a). One may observe that there are many points on the x-axis, which indicates that in many cases $I_a(t)$ will not cause the output to change at all. Other points lie roughly on a line with a slope $k\approx 1$, indicating that the two variables are highly correlated. We then plot the scatter points separated by the value of somatic membrane potential $u(t)$ at each sampling time $t$ in Figure \ref{fig:apical} (b) and (c). Clearly, the correlation between the two axes is higher when $u$ is close to the threshold $\vartheta$. In Figure \ref{fig:apical} (d), we plot the correlation slope $k$ versus $u$ for a set of different norm-ratio between apical/basal inputs $||I_a||_2/||I_b||_2$. The shaded area shows the variance of the slope in 1000 repeated rounds. Interestingly, the shape of these lines is consistent with the surrogate derivative \cite{zenke2018superspike,shrestha2018slayer,wu2018spatio}, which peaks when $u=\vartheta$ and decrease gradually as $u$ moves away from $\vartheta$. 

\subsubsection{Neuron Internal Error Propagation}
\begin{figure}[t]
    \centering
    \includegraphics[width=0.95\textwidth]{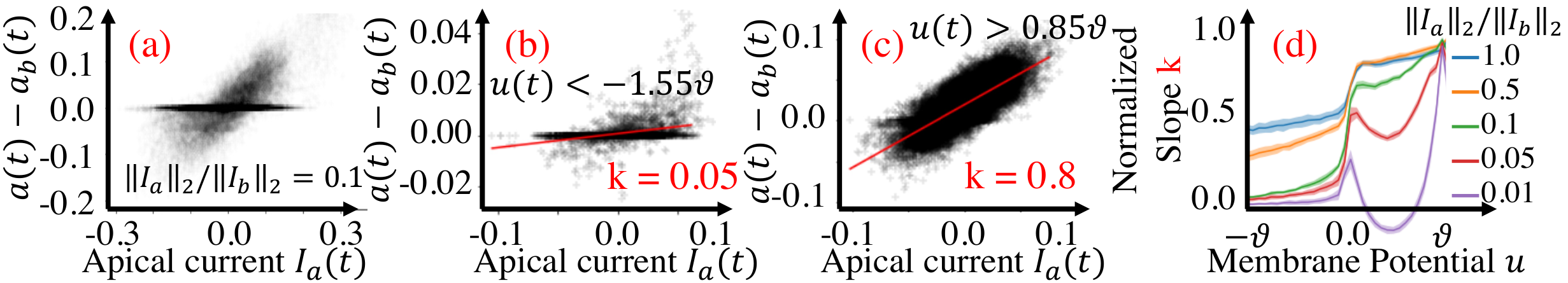}
    \caption{(a) Scatter plot between apical dendrites current and change on PSC. (b) membrane potential $u(t)<-1.55\vartheta$. The red line is fitted by linear regression, which shows small slope $k=0.05$. (c) The correlation when $u(t)>0.85\vartheta$. $I_a$ strongly modulates the PSC, where $k=0.8$. (d) Slope v.s. membrane potential among a set of apical/basal norm-ratio.}
    \label{fig:apical}
\end{figure}

We first suppose that the current flowing through the apical dendritic compartment encodes the negative partial derivative of loss w.r.t the neuron's PSC: $I_a = -\partial L/\partial a$. Statistically, we argue that the relationship between small-signal apical dendrites current and a neuron's PSC satisfies the linear addition assumption, where the slope is higher as membrane potential moves closer to the threshold. 
Since the long term synaptic plasticity emerges from integration effect on a relatively long time-window, where noise is averaged out, we ignore the noisy correlation as in Figure \ref{fig:apical}, and use a surrogate function $\sigma'(u) = 1/(1+|u-1|)^2$ to describe the small-signal relationship to accelerate the training. When verify the effectiveness of our framework, using spiking based backward model is unnecessary as it greatly increased the simulation time. The total backward PSC of a Pyr neuron is modeled as: $a(t) = a_b(t) + \sigma'(u(t))I_a(t)$.
For inhibitory PV cells, the output PSCs are negatively correlated to their  apical dendrites current inputs, so the total backward PSC of a PV cell is modeled as:  $a(t) = a_b(t) - \sigma'(u(t))I_a(t)$. 
The two rules are unified as $a(t) = a_b(t) + e(t)$, where $e$ is:
\begin{equation}
    e = \left\{
    \begin{array}{lr}
    \sigma'(u) I_a = -\sigma'(u)({\partial L}/{\partial a}),& {\rm Pyr~cells} \\
    -\sigma'(u) I_a = \sigma'(u)({\partial L}/{\partial a}),& {\rm PV~cells}
    \end{array}\right.
    \label{eq:error}
\end{equation}
The small error signal locates in the apical dendrites continuously modulates the neuron's backward PSC without effect the forward process. Therefore, if we can disentangle at all time $e(t)=a(t)-a_b(t)$ (shown in Section \ref{sec:exc_back} \& Section \ref{sec:inh_back}), we can achieve error backpropagation without separated forward and backward phases (\textbf{IS.3}).

\subsection{Forward Path}

The connections from Pyr and PV cells in one layer to the basal dendrites of Pyr and PV cells in their next layer build up the forward path (\textbf{IS.5}). Consider a fully connect layer, where we  express Pyr and PV neurons' PSCs ($\boldsymbol{a}_+$ and $\boldsymbol{a}_-$) and synaptic weights ($\boldsymbol{\rm W}_{\rm Pyr}$ and $\boldsymbol{\rm W}_{\rm PV}$) separately. The forward path is expressed as:
\begin{equation}
    \boldsymbol{I}^{(l+1)} = \boldsymbol{\rm W}_{\rm Pyr}^{(l+1)} \boldsymbol{a}_+^{(l)} + \boldsymbol{\rm W}_{\rm PV}^{(l+1)} \boldsymbol{a}_-^{(l)}
\label{eq:forward}
\end{equation}

Generally, neurons that involved in the forward path also needs to propagate errors backward, which requires separate compartments to store both forward activation signals and backward error signals at the same time. (\textbf{NE.2}) suggests that Pyr and PV neurons have such special structure. Meanwhile, neurons in the forward path synapse onto neurons with the same type in its next layer. According to \textbf{NE.4}: PV cells strongly inhibit one another. Yet SOM cells are rarely observed inhibiting each others \cite{pfeffer2013inhibition}. Therefore SOM cells are not suitable to participate in the forward path directly. Furthermore, \textbf{NE.5}: SOM cells' axons are known mainly connect to Pyr cells' apical dendrites, and PV cells' axons are known to mainly connect to Pyr cells' somata or basal dendrites \cite{chen2015subtype}. So we use basal dendrites to receives the forward signal, and apical dendrites to store the backward signal. This setting is further supported by \textbf{NE.6}: Apical dendrites of PV neurons were observed to form bundles and co-localize with apical dendritic bundles of Pyr neurons in retrosplenial cortex \cite{ichinohe2002parvalbumin}. Such co-localization potentially allows Pyr and PV neurons to share the same error information.

Early works failed to follow the Dale's Principle \cite{sacramento2018dendritic, zenke2018superspike}. They did not include inhibitory neurons in the forward path, but gave synaptic weights the flexibility to switch between positive and negative. Besides, similar approaches have difficulty explaining how to realize bio-plausible sign-concordant feedback alignment \cite{moskovitz2018feedback}. Under our framework, forward backwards weights are naturally sign-concordant for all weights are positive (refer to Section \ref{sec:background_snn}). Meanwhile, the all positive weights setting has not hinder the training performance for we introduced inhibitory neurons in the forward path.

\subsection{Excitatory Backward Path}
\label{sec:exc_back}
\begin{figure}[t]
    \centering
    \includegraphics[width=0.95\textwidth]{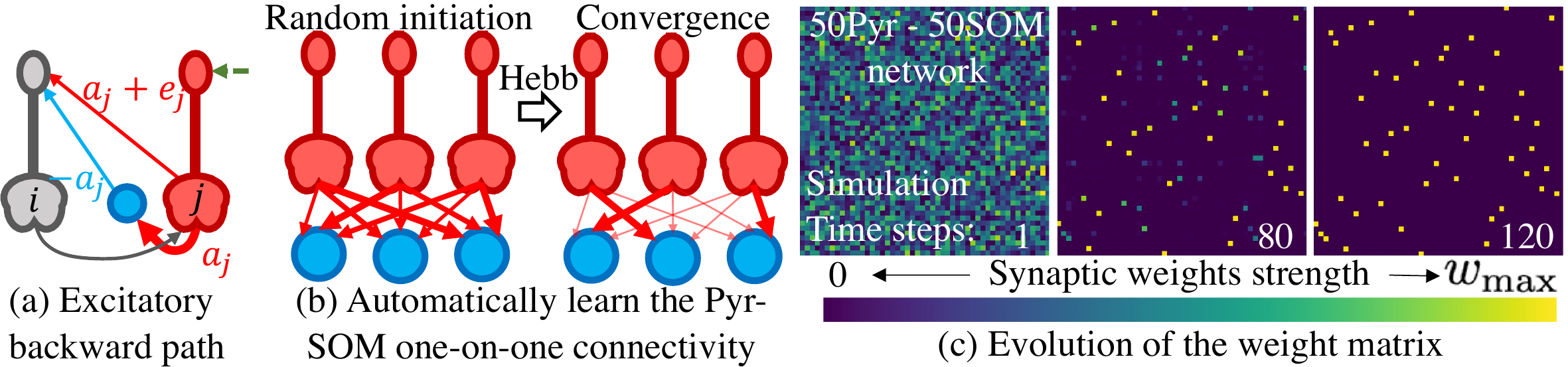}
    \caption{Excitatory backward microcircuit and the assembly competition mechanism}
    \label{fig:exc_back}
\end{figure}

As in Figure \ref{fig:exc_back} (a), the gray colored neuron represents a nonspecified neuron from the Pyr neuron's previous layer. Computationally, a Pyr neuron that receives inputs from its previous layer, needs to propagate its error information backwards. Yet the spike-based backward PSCs $a(t) = a_b(t) + e(t)$ has a large base PSC current $a_b(t)$. Following previous work \cite{sacramento2018dendritic}, we also introduce an additional inhibitory cell to cancel out $a_b(t)$. As observed in \textbf{NE.7}: there exists assembly competition mechanism \cite{silberberg2008polysynaptic}, which drives SOM to follow the most active Pyr \cite{tremblay2016gabaergic}, we design our excitatory backward microcircuit with one additional SOM neuron, which receives a one-on-one input from its paired Pyr neuron, and synapse onto the apical dendrites (following \textbf{NE.5} and \textbf{NE.6}) of neurons in the previous layer. In other words, every neuron, no matter Pyr or PV, that synapses onto a Pyr neuron's basal dendrites receives two connections synapse back onto its apical dendrites. One from the Pyr neuron, the other from the paired SOM neuron (\textbf{IS.5}). Name the Pyr backward weights in layer $(l+1)$ as $\boldsymbol{\rm W}_{\overleftarrow{\rm Pyr}}^{(l+1)}$, and the Pyr backward PSCs as $\boldsymbol{a}_{\overleftarrow{+}}^{(l+1)}$, The pyramidal cells' backward currents are simply formulated as $\boldsymbol{\rm W}_{\overleftarrow{\rm Pyr}}^{(l+1)} \boldsymbol{a}_{\overleftarrow{+}}^{(l+1)}$. 
Name the backward weights matrix of SOM cells in layer $(l+1)$ as $\boldsymbol{\rm W}_{\overleftarrow{\rm SOM}}^{(l+1)}$, which shares the same shape with $\boldsymbol{\rm W}_{\overleftarrow{\rm Pyr}}^{(l+1)}$. For excitatory (+) backward path, the top down current is:
\begin{equation}
\begin{aligned}
    \boldsymbol{I}_{a+}^{(l)}&=\boldsymbol{\rm W}_{\overleftarrow{\rm Pyr}}^{(l+1)}\cdot\boldsymbol{a}_{\overleftarrow{+}}^{(l+1)} + \boldsymbol{\rm W}_{\overleftarrow{\rm SOM}}^{(l+1)} \cdot\boldsymbol{a}_{\rm SOM}^{(l+1)}\\
    &= \boldsymbol{\rm W}_{\overleftarrow{\rm Pyr}}^{(l+1)}\cdot(\boldsymbol{a}_{b+}^{(l+1)} + \boldsymbol{e}_{+}^{(l+1)}) + \boldsymbol{\rm W}_{\overleftarrow{\rm SOM}}^{(l+1)} \cdot (-\boldsymbol{a}_{b+}^{(l+1)}).
    \label{eq:exc_back}
\end{aligned}
\end{equation}
Ideally, only the pure error signal $\boldsymbol{e}_{+}$ is propagated backwards, and $\boldsymbol{\rm W}_{\overleftarrow{\rm Pyr}}\cdot\boldsymbol{a}_{b+}$ is canceled out by $\boldsymbol{\rm W}_{\overleftarrow{\rm SOM}} \cdot (-\boldsymbol{a}_{b+})$, which implies $\boldsymbol{\rm W}_{\overleftarrow{\rm Pyr}} = \boldsymbol{\rm W}_{\overleftarrow{\rm SOM}}:= \boldsymbol{B}_+$. This alignment is guaranteed by the anti-Hebbian learning, which we will discuss in Section \ref{sec:learning}.

The only hypothesis we make is that the sole connection needs to be strong enough (greater than SOM cells' threshold) so that each excitatory spike fired by Pyr cells can induce an inhibitory SOM spike. Fortunately, \textbf{NE.8}: SOM is well-known for its low threshold, and the ability to fire temporal accurate inhibitory spikes\cite{urban2016somatostatin}. Our SOM neuron model differs from the previous work: We present a SOM model with single compartment (respecting NE.3), while Sacramento\el \cite{sacramento2018dendritic} model a SOM cell with double compartments (the additional one is used to store a "nudging signal" from its paired Pyr neuron). Therefore, we adjust synapses that connect to SOM cells error-freely.

The error-free assembly competition mechanism is realized through Hebbian learning, which can be informally described as: "Cells that fire together, wire together" \cite{hebb1949organization}. Yet cells that wires together also tend to fire together more often. Due to the existence of such a positive feedback loop, one needs to introduce additional tricks to stabilize the learning process - such as weights normalization \cite{ferre2018unsupervised}, or weights clamping \cite{diehl2015unsupervised, kheradpisheh2018stdp}. Here, we find that with a proper normalization, the one-on-one connectivity can be formed easily.

Figure \ref{fig:exc_back} (b) exhibits our simulation goal. Name the synaptic weights from Pyr cell to SOM cell within the same layer $(l)$ as $\boldsymbol{\rm W}_{\rm PyS}^{(l)}$. We visualized $\boldsymbol{\rm W}_{\rm PyS}^{(l)}$ in Figure \ref{fig:exc_back} (c): The weight matrix gradually forms one-on-one connectivity pattern from pure random initiation.
The Hebbian rule we applied here is simple. We stimulate Pyr cells randomly, and records the firing behavior of SOM cells. The synaptic weights of cells that fires together are strengthened, otherwise weakened.  

The normalization we apply is to keep the summation along both rows and columns of $\boldsymbol{\rm W}_{\rm PyS}^{(l)}$ as a constant. Consider the total area of SOM cell's dendrites are limited, forming more synaptic connections with a Pyr cell requires weaken other Pyr cells' connections. Similarly, if a Pyr cell wish to form more axon connections to a SOM cell, a price of weakening other axon connections are paid. As demonstrated in the last sub-figure of Figure \ref{fig:exc_back} (c), each SOM cell successfully find their paired Pyr cell to follow with.

\subsection{Inhibitory Backward Path}
\label{sec:inh_back}

\begin{figure}[t]
    \centering
    \includegraphics[width=0.99\textwidth]{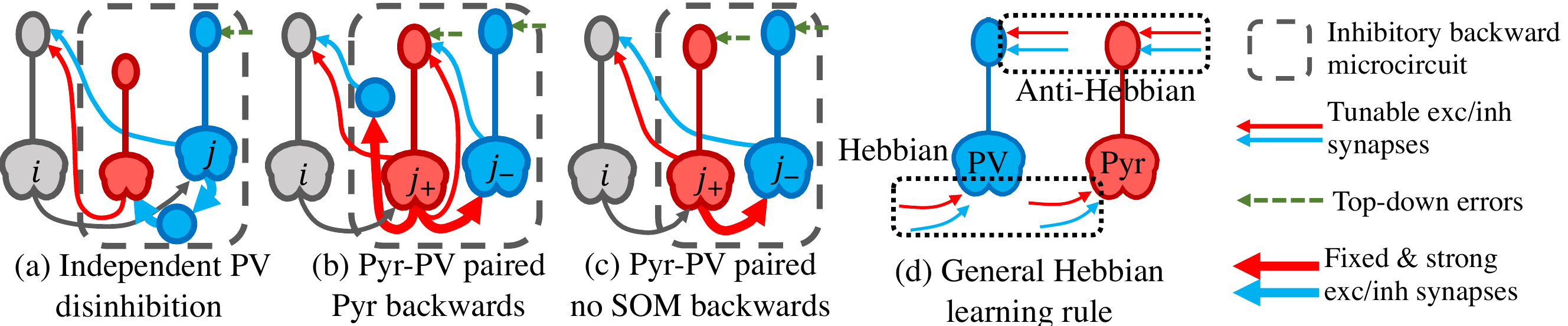}
    \caption{Inhibitory backward microcircuits}
    \label{fig:inh_back}
\end{figure}

We find several hints regarding how may PV cells backpropagates their errors backwards, and conclude them as following (\textbf{IS.5}). One of the first microcircuits we found plausible is named as "disinhibition", where an excitatory spike is generated through relieving the inhibitory inputs from an excitatory neuron. As observed in \textbf{NE.9}: PV and SOM cells synapse onto each others and consequently have a disinhibition effect \cite{tremblay2016gabaergic}. Here we surmise that the disinhibition connection PV to SOM might related to the inhibitory error backpropagation process. As shown in Figure \ref{fig:inh_back} (a), the PV cell inhibits a SOM cell from emitting inhibitory PSCs to a Pyr cell, and thus the Pyr cell is able to emit an excitatory spike backwards. The Pyr cell used here does not involved in the forward path, and does not receive current inputs from its apical dendrites. Additional bias currents inputs targeting the SOM and Pyr cells may required in this microcircuit. Name the backward weights matrix of PV cells in layer $(l+1)$ as $\boldsymbol{\rm W}_{\overleftarrow{\rm PV}}^{(l+1)}$, and the one of their paired Pyr cells as $\boldsymbol{\rm W}_{\overleftarrow{\rm Pyr_{(PV)}}}^{(l+1)}$. 
For inhibitory (-) backward path, the top down current is:
\begin{equation}
\begin{aligned}
    \boldsymbol{I}_{a-}^{(l)}&=\boldsymbol{\rm W}_{\overleftarrow{\rm PV}}^{(l+1)}\cdot\boldsymbol{a}_{\overleftarrow{-}}^{(l+1)} + \boldsymbol{\rm W}_{\overleftarrow{\rm Pyr_{(PV)}}}^{(l+1)} \cdot\boldsymbol{a}_{\rm Pyr_{(PV)}}^{(l+1)}\\
    &= \boldsymbol{\rm W}_{\overleftarrow{\rm PV}}^{(l+1)}\cdot(\boldsymbol{a}_{b-}^{(l+1)} + \boldsymbol{e}_{-}^{(l+1)}) + \boldsymbol{\rm W}_{\overleftarrow{\rm Pyr_{(PV)}}}^{(l+1)} \cdot (-\boldsymbol{a}_{b-}^{(l+1)}).
\end{aligned}
\end{equation}
The inhibitory backward path also following the same spirit as (\ref{eq:exc_back}), and have an equivalent backward weights $\boldsymbol{\rm W}_{\overleftarrow{\rm PV}}^{(l+1)} = \boldsymbol{\rm W}_{\overleftarrow{\rm Pyr_{(PV)}}}^{(l+1)} :=\boldsymbol{B}_-$ when two weights aligned. The overall backward weights $\boldsymbol{B}$ is the concatenation of $\boldsymbol{B}_+$ and $\boldsymbol{B}_-$.

The second hypothetical microcircuit is inspired by \textbf{NE.10}: Pyr cells forms autapses (self-connect), which represents a key circuit element in the neocortex \cite{yin2018autapses}.
Though more complicated corporations could form within a group of Pyr-PV cells, here we only discuss a simple one-on-one pairing within the same layer from Pyr cells to PV cells. The strong connection synchronize a paired cells' firing behavior. As shown in Figure \ref{fig:inh_back} (b), the PV cell only receives basal inputs from its paired Pyr cell, and accumulates its error back onto its paired Pyr cell's apical dendrites.The Pyr cell will propagate both PV cells' error information and errors of itself backwards together, with the help of its paired SOM cell as in Figure \ref{fig:exc_back} (a). Please refer to the appendix for a detailed explanation.

Further, we present our third microcircuit. As shown in Figure \ref{fig:inh_back} (c), paired Pyr-PV cells can propagate their errors backwards without the help of SOM cells. The only backward synapses are send by a pair of Pyr and PV cells to apical dendrites of neurons in their previous layer.


We modestly suspect that all the three microcircuits may co-exist in a layer to propagate PV cells error backwards. Worth to note, the second and third microcircuits only consider one-on-one cooperation between Pyr cells and PV cells within a same layer, more complex cooperation within a group of Pyr, PV, and SOM cells within a same layer may exist, and the rules to form such connections appears to be intriguing for future works.

\section{Learning Rules}
\label{sec:learning}
\subsection{General Hebbian Learning Rule}
The most fundamental version of Hebbian rule calculates the correlation between pre- and post-synaptic signals, the higher the correlation, the more the synapse is strengthened. In our work, we adopt this simplest version and define the Hebbian rule as the dot product between the pre- and the postsynaptic signals $\boldsymbol{x}_{\rm pre}$ and $\boldsymbol{y}_{\rm post}$ scaled by a learning rate $\eta$:
\begin{equation}
    {\rm Hebb({x}_{\rm pre},{y}_{\rm post})}=\eta \int {x}_{\rm pre}(t){y}_{\rm post}(t) dt
\end{equation}
We further define anti-Hebbian rule as the negative version of Hebbian rule.
Our proposed general Hebbian learning rule is inspired by \textbf{NE.11}: Pyr neurons have different plasticity rules between apical dendrites and somata. When an outside simulation is applied, apical dendrites synapses are observed to have long-term-potentiation (LTP), yet somata synapses are observed to have long-term-depression (LTD) \cite{udakis2020interneuron}. (\textbf{NE.11}) indicates that the plasticity rule between apical dendrites and somata might be opposite. Coincidentally, such a rule fits our computational needs well (solving \textbf{IS.3}). Our proposed learning rule as shown in Figure \ref{fig:inh_back} (d) applies:
\begin{equation}
    \Delta w_{ij} \propto  \left\{
    \begin{array}{lr}
    {\rm Hebb}(a_j, e_i),& {\rm Basal~dendrites} \\
    -{\rm Hebb}(a_j, {I_a}_i),& {\rm Apical~dendrites}
    \end{array}\right.
\end{equation}
In our framework, a positive postsynaptic error signal $e$ requires a neuron to fire more spikes, which means to strengthen the excitatory input and weaken the inhibitory input connected to basal dendrites. Therefore the correlation between error and input PSCs reflects the weight adjusting directions, which yield Hebbian rule.

Yet for synapses connect to apical dendrites, imbalance excitatory-inhibitory backward weights leads to an imbalance bias on the postsynaptic apical dendrites $I_a$, where even the same excitatory-inhibitory firing patterns cannot canceling each others perfectly. Therefore, the adjusting goal for feedback weights is to minimize apical dendrites' signals. An over-excitatory apical dendrites requires excitatory Pyr cells' backward weights to decrease, and PV's backward weights to increase. The anti-Hebbian rule pushes synaptic weights that transmit highly correlated excitatory and inhibitory signals to have the same strength. So they can cancel each other, minimize the norm of apical dendrites signal, and leave only clean error information.
Importantly, the resulting equivalent weights matrix $\boldsymbol{B}$ usually has different magnitude with their corresponding forward weights matrix $\boldsymbol{W}^T$, which is in favour of the idea of random backpropagation.

\subsection{Equivalence to SFA}
Suppose that the current flowing through the apical dendrites of the output-layer neurons encodes error $(-dL/d\boldsymbol{a})$. Then in our framework, the current $\boldsymbol{I}_a$ flowing through the apical dendrites in the layers other than the output layers encodes an \textbf{approximated} error signal. The approximation comes from two points: \textbf{1)} The imperfect alignment of the excitatory and inhibitory backward weights as described in Sec \ref{sec:exc_back} and \ref{sec:inh_back}.
\textbf{2)} Even assume all backward weights are perfectly converged (aligned with their corresponding weights). The converged values $\boldsymbol{B}$ are not equal to the transpose of the forward weight $\boldsymbol{W}^T$. The point \textbf{1)} comes from our framework's biological constrains, and the point \textbf{2)} comes from the standard SFA (Sec 2.3). Since following Dale's law naturally lead to the following of the sign-concordant property: All weights in $\boldsymbol{W}$ and $\boldsymbol{B}$ are positive in our framework so they must share the same sign, so we do not need to route the sign of the forward weight $\boldsymbol{W}$(concatenation of $\boldsymbol{W}_{\rm Pyr}$ and $\boldsymbol{W}_{PV}$) to $\boldsymbol{B}$ as previous SFA works\cite{liao2016important, moskovitz2018feedback}. Therefore, when we assume perfect alignment, the framework propagate errors across layers with \textbf{2)} only, which is equivalent to SFA.

\section{Experimental results}
\label{sec:exp}

The proposed framework is compared with other BP-based rules on two widely used real-world datasets: MNIST \cite{lecun1998mnist} and CIFAR-10 \cite{krizhevsky2009learning}. We use the fixed-step first-order forward Euler method to discretize continuous membrane voltage updates over a set of discrete time steps. The results are concluded in table \ref{tab:MNIST_CIFAR10}. Our method gains comparable performance as compared to BP-based works.  Importantly to know, the remarkable performance is gained upon bio-plausible settings.
\begin{table}[h]
    \centering
    \caption{Performances comparison on the MNIST and CIFAR10}    
    \begin{tabular}{c|c|c|c|c}
        \hline
         & \multicolumn{2}{c|}{MNIST} & \multicolumn{2}{c}{CIFAR10} \\
        \hline
        Method  & \#Discret time steps & BestAcc & \#Steps & BestAcc\\
        \hline
        SLAYER \cite{shrestha2018slayer} & 300 & 99.41\%  & null & null\\
        TSSL-BP \cite{zhang2020temporal}  & 5 & 99.53\%  & 5 & 89.22\%\\
        NA \cite{yang2021backpropagated} & 5 & \textbf{99.69\%} & 5 & \textbf{91.76\%} \\
        Ours & 5 & 99.64\%  & 5 & 86.88\%\\
        \hline
        \multicolumn{5}{l}{MNIST: 15C5-P2-40C5-P2-300}\\
        \multicolumn{5}{l}{CIFAR: 96C3-256C3-P2-384C3-P2-384C3-256C3-1024-1024}
    \end{tabular}
    \label{tab:MNIST_CIFAR10}
\end{table}

We made some bio-implausible trade-offs when benchmarking with BP-based methods, including: 1) Assume clear targets and errors are provided for these two supervised learning tasks. 2) The standard computer vision convolution layer (\#channel-C-kernel size) and pooling layer (P-kernel size) are applied. 3) The weights of the input layer are allowed to be both positive and negative to handle the normalized input signals. 4) The Adam optimizer \cite{loshchilov2017decoupled} is used without a clear biologically explanation. 5) We assume perfect excitatory inhibitory backward alignment. 6) The layer-wise architecture is left biologically undefined. Please refer to the appendix for more experimental results.


\section{Conclusion}
In this work, we construct a novel framework to implement bio-plausible random error backpropagation. The framework consists of three key parts: new neuron models, new microcircuits, and supporting local learning rules. Guided by Dale's Principle, we construct our architecture with excitatory Pyr neurons, inhibitory PV and SOM neurons. Instead of using a rate-based model, we employ the spike-timing-based LIF model for all types of neurons, and we further provide a new model of the apical dendrites internal dynamics, which addresses the problem of propagating continuous errors through discrete spikes. Motivated by the fact that SOM cells have spatially constrained dendrites, we implement the assembly competition mechanism observed in brain, which forms one-on-one Pyr-SOM pairs automatically and help Pyr neurons to propagate errors backwards. Meanwhile, we propose three hypothetical microcircuits to address the issues of error propagation in inhibitory PV neurons. Based on our architecture, we propose to apply opposite local learning rules on two sets of synapses. Hebbian updates are used to adjust the weights of synapses connected to basal dendrites, while anti-Hebbian updates are used to adjust the weights of synapses connected to apical dendrites. Our framework is evaluated on MNIST and CIFAR10 datasets and obtains BP-comparable accuracy. 
In the future, we would like to explore the effect of forming more dependencies between neurons within the same layer, and possibly come up with more general learning rules that can automatically form microcircuits that can support learning. We hope that this work will foster connections between deep learning and neuroscience communities.

\bibliographystyle{unsrt}
\bibliography{reference}

\appendix
\newpage
\section{Additional experimental results}

\subsection{Anti-Hebbian rule}
\label{sec:anti_exp}
We designed an experiment to test the convergence ability of our anti-Hebbian rule with different energy of the backward error signals. We randomly generate 50 pairs spike trains, where each pair have exactly opposite values: excitatory spike = 1, while inhibitory spike = -1. The random spikes followed the Bernoulli distribution with $P({\rm fire})=0.02$ at each simulation time-steps. There are 50 receiving apical dendrites, so both the synapses weights $\boldsymbol{\rm W}_{\overleftarrow{\rm Pyr}}$ and $\boldsymbol{\rm W}_{\overleftarrow{\rm SOM}}$ have shape (50, 50). The total apical dendrites inputs are:
$\boldsymbol{I}_a[t] =\boldsymbol{\rm W}_{\overleftarrow{\rm Pyr}} \boldsymbol{a}_{\overleftarrow{+}}[t] + \boldsymbol{\rm W}_{\overleftarrow{\rm SOM}}\boldsymbol{a}_{\overleftarrow{-}}[t].$
Apply the anti-Hebbian rule:
\begin{equation}
\begin{aligned}
&\Delta \boldsymbol{\rm W}_{\overleftarrow{\rm Pyr}}[t] = -\eta \cdot \boldsymbol{a}_{\overleftarrow{+}}[t] \otimes \boldsymbol{I}_a[t]\\
&\Delta \boldsymbol{\rm W}_{\overleftarrow{\rm SOM}}[t] = -\eta \cdot \boldsymbol{a}_{\overleftarrow{-}}[t] \otimes \boldsymbol{I}_a[t]
\end{aligned}
\end{equation}
where $\otimes$ is the outer product. Later, we add an error signal onto $\boldsymbol{a}_{\overleftarrow{+}}$. Modeling the backward error signals as Gaussian noise with standard deviation varies from 0.01 to 1.0.
The results are shown in Figure \ref{fig:exp} (a). The experiment runs 50 times, and the shaded area is the standard deviation. The results exhibit that $\boldsymbol{\rm W}_{\overleftarrow{\rm Pyr}}-\boldsymbol{\rm W}_{\overleftarrow{\rm SOM}}$ converges to near zero most of the time, which proves that the anti-Hebbian rule can tolerant certain strength of error signals.

\subsection{Separate phase to align excitatory and inhibitory backward weights}
Indeed, a separate phase to align backward weights benefits the training, and we assume two weights are equal when benchmarking with other BP-based methods, but this is not required. We add an experiment to compare the learning behavior with/without a separate phase as shown in Figure \ref{fig:exp} (b). Without a separate phase, the final accuracy drops a little bit due to the imperfect alignment of the excitatory and inhibitory backward weights as described in Sec \ref{sec:exc_back} and \ref{sec:inh_back}.

\subsection{Feedback weight angle alignment}

\begin{figure}[ht]
    \centering
    \includegraphics[width=0.9\textwidth]{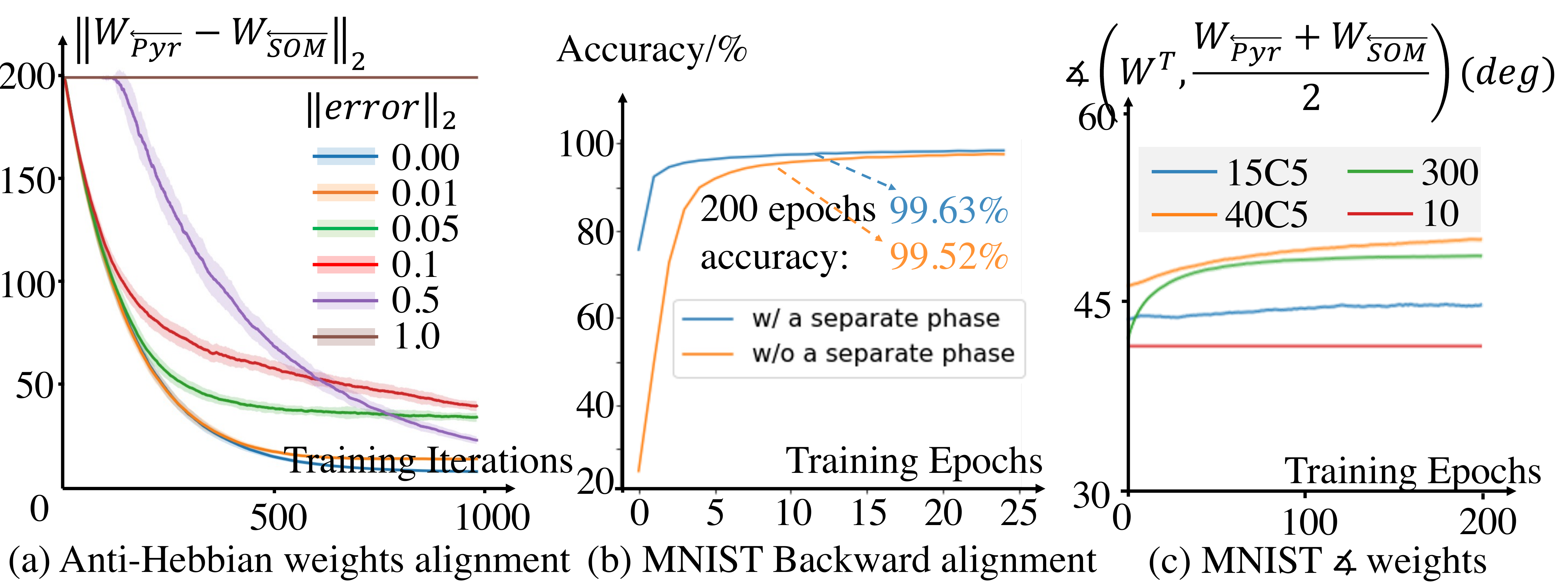}
    \caption{Additional experiments}
    \label{fig:exp}
\end{figure}

We test the weight angle between our random backward weights and the transpose of the forward weights on MNIST. Since our framework satisfy sign-concordant automatically, the angle are always near 45\textdegree as shown in Figure \ref{fig:exp} (c).

\section{Detailed backward microcircuits}
\subsection{Excitatory backward microcircuit}

We propose an excitatory backward path mediated by Pyr neurons paired each with one SOM neuron, whose role is to propagate the error signals (via $\boldsymbol{\rm W}_{\overleftarrow{\rm Pyr}}$) and cancel out self-generated top-down input to the neurons in the previous layer (via $\boldsymbol{\rm W}_{\overleftarrow{\rm SOM}}$) and this way leave the backward synaptic currents encoding a pure error signal (as Sacramento\el \cite{sacramento2018dendritic} and Yang\el \cite{yang2021bioleaf}). Furthermore, the pairing of each Pyr neuron with its SOM neuron is achieved through an assembly competition mechanism. The role of this building block is to emulate the backpropagation of error signals through Pyr neurons.

\begin{figure}[ht]
    \centering
    \includegraphics[width=0.9\textwidth]{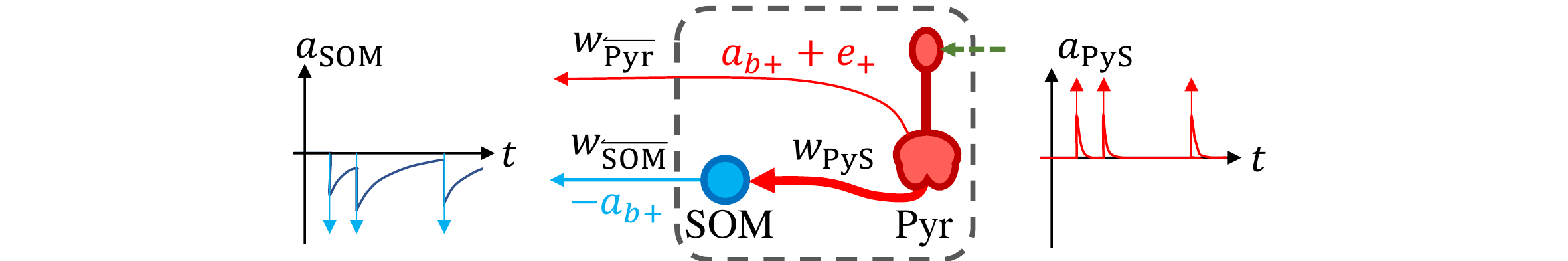}
    \caption{Excitatory backward microcircuit}
    \label{fig:exc_back_micro}
\end{figure}

As shown in the above Figure \ref{fig:exc_back_micro}, the excitatory backward microcircuit has an auxiliary SOM neuron, which receives a one-on-one input current $a_{{\rm PyS}_i}^{(l)} = a_{+_i}^{(l)}$ from the Pyr neuron in the same layer through the synapse $w_{{\rm PyS}_i}^{(l)}$:
\begin{equation}
    I_{{\rm SOM}_i}^{(l)}(t) = w_{{\rm PyS}_i}^{(l)}\cdot a_{{\rm PyS}_i}^{(l)}(t)
\end{equation}

In the microcircuit, a SOM neuron needs to fire a spike after receiving a spike, which works when setting the exponential decaying time constant $\tau_s$ of the synapse $w_{\rm PyS}$ to be a small value. Short time constant leads to a fast exponential decaying shape of PSC. Together with a proper short refractory time window, the PSC induced by a single spike cannot induce multiple spikes of a SOM neuron.

Experimentally, we discretize the continuous-time waveforms over a set of discrete time-steps using the fixed-step first-order forward Euler method. The general dynamics of a neuron's membrane potential ${u}_i$ is described by the following equations:
\begin{equation}
{u}_i[t+0.5] = \left(1-\frac{1}{\tau_m}\right){u}_i[t] +\sum_{j} w_{ij}{a}_j[t+1].
\label{eq:mem}
\end{equation}
\begin{equation}
    {s}_i[t] = H\left({u}_i[t+0.5]-\vartheta\right),
    \label{eq:spike}
\end{equation}
\begin{equation}
    {u}_i[t+1] = {u}_i[t+0.5] \left(1-{s}_i[t]\right)
    \label{eq:reset_zero}
\end{equation}

When ${u}_i$ reaches the threshold $\vartheta$, a spike is generated through (\ref{eq:spike}), and the membrane potential is reset to zero through (\ref{eq:reset_zero}). The dynamics of synapses is:
\begin{equation}
    {a}_i[t+1]=\left(1-\frac{1}{\tau_s}\right){a}_i[t]+\left(\frac{1}{\tau_s}\right){s}_i[t+1].
    \label{eq:a_time}
\end{equation}
For the excitatory backward microcircuit, we set $\tau_s = 1$ for synapse $w_{\rm PyS}$. Then each Pyr spike induce a PSC exist on a single time-step only with value $a_{\rm PyS}=1$. We set the converged maximum value of $w_{\rm PyS}$ as one, so the overall input current $\left(w_{\rm PyS}\cdot{a}_{\rm PyS}\right) = 1$. Therefore, the threshold of SOM neurons should locate in the range of $0<\vartheta < 1$. This guarantees each excitatory Pyr spike induces an inhibitory SOM spike. We set $\vartheta=0.5$ in our simulation. 

The membrane potential of a SOM cell is always zero, except when it received an single time-step excitatory input: The SOM cell has ${u}_i[t+0.5] = 1$, which leads to a spike firing ${s}_i[t]=1$, and ${u}_i[t+1] = {u}_i[t+0.5] \left(1-{s}_i[t]\right) = 0$ back to zero again. In the whole process, the time constant of membrane potentials ${\tau_m}$ of SOM cells does not affect the simulation, and its value is free to choose.

\begin{figure}[ht]
    \centering
    \includegraphics[width=0.9\textwidth]{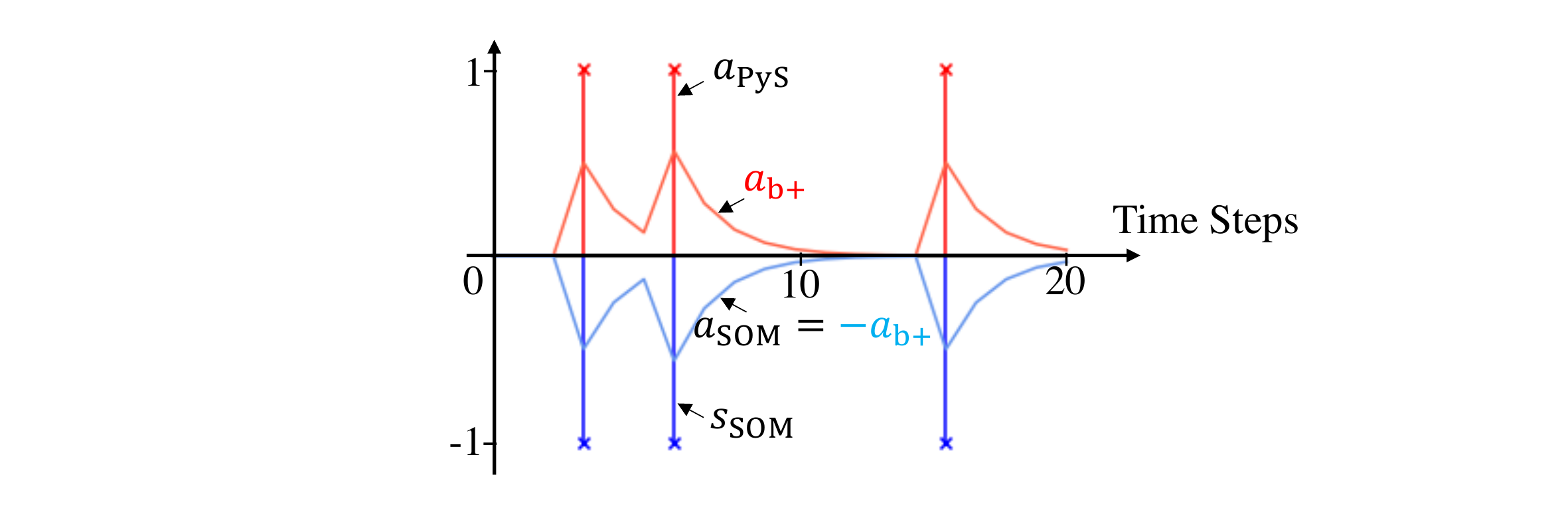}
    \caption{Excitatory backward microcircuit discrete time simulation. $\# {\rm Time~steps} = 20$, $\tau_s=2$}
    \label{fig:exc_back_micro_1}
\end{figure}

For the output synapses of SOM neurons, we set $\tau_s$ of $w_{\overleftarrow{\rm SOM}}$ equal to the $\tau_s$ of its paired Pyr neuron's backward synapses, so the inhibitory and excitatory PSCs can canceling each others perfectly when $w_{\overleftarrow{\rm Pyr}}=w_{\overleftarrow{\rm SOM}}$. An example waveform of the discrete time simulation is shown in Figure \ref{fig:exc_back_micro_1}.

\subsection{Inhibitory backward microcircuits}
All the three inhibitory microcircuits are heuristically designed based on neuroscience observations. Yet we do not provide a learning rule explaining how these connections formed. Our goal in this section is to make our microcircuits' structure clear. 

Like Pyr neurons, PV neurons also have self-generated base PSC signal $a_{b-}$ that need to be cancel out by a paired excitatory PSC $-a_{b-}$. We achieve this through three possible microcircuits, which we will demonstrate them one by one in this section.

\subsubsection{Inhibitory backward microcircuit 1}

The first inhibitory microcircuit is inspired from the disinhibition connections as shown in the Figure \ref{fig:inh_back_micro} below:
\begin{figure}[ht]
    \centering
    \includegraphics[width=0.9\textwidth]{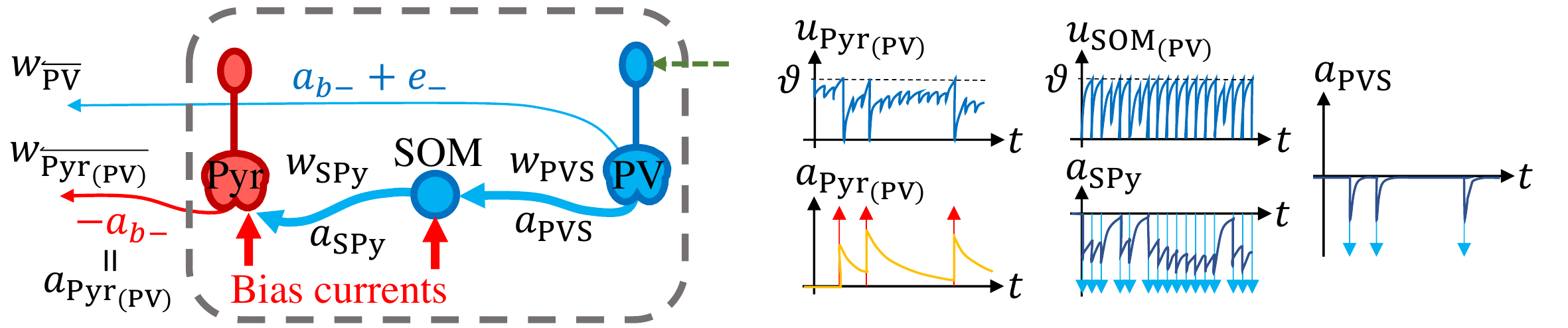}
    \caption{Inhibitory backward microcircuit 1}
    \label{fig:inh_back_micro}
\end{figure}
Each PV neuron is paired with one auxiliary SOM neuron and one auxiliary Pyr neuron. The input current of the SOM neuron $i$ in layer $(l)$ is:
\begin{equation}
    I^{(l)}_{\rm SOM_{{(PV)}_i}}(t) = w^{(l)}_{{\rm PVS}_i}\cdot a^{(l)}_{{\rm PVS}_i}(t) + I_{bias{\rm (SOM)}}
\end{equation}
The SOM interneuron is strongly excited by an external excitatory bias current $I_{bias{\rm (SOM)}}$, so it keeps emitting inhibitory spikes to a Pyr neuron except for when the SOM cell itself is inhibited by the PV cell. A Pyr neuron in this microcircuit receives input current:
\begin{equation}
    I^{(l)}_{\rm Pyr_{{(PV)}_i}}(t) = w^{(l)}_{{\rm SPy}_i}\cdot a^{(l)}_{{\rm SPy}_i}(t) + I_{bias{\rm (Pyr)}}
\end{equation}
We choose the value of $w^{(l)}_{{\rm SPy}_i}$ and $I_{bias{\rm (Pyr)}}$ so that the firing behavior of a pair of SOM and Pyr cells are also contrary.

Once an inhibitory spike is generated from the PV cell, it offset the bias current $I_{bias{\rm (SOM)}}$, so the SOM cell stops firing temporarily, which alleviates the Pyr cell from inhibition. The Pyr cell would then generate an excitatory output spike. This effect is known as disinhibition, which is prevalent in the brain \cite{tremblay2016gabaergic}.

\begin{figure}[ht]
    \centering
    \includegraphics[width=0.99\textwidth]{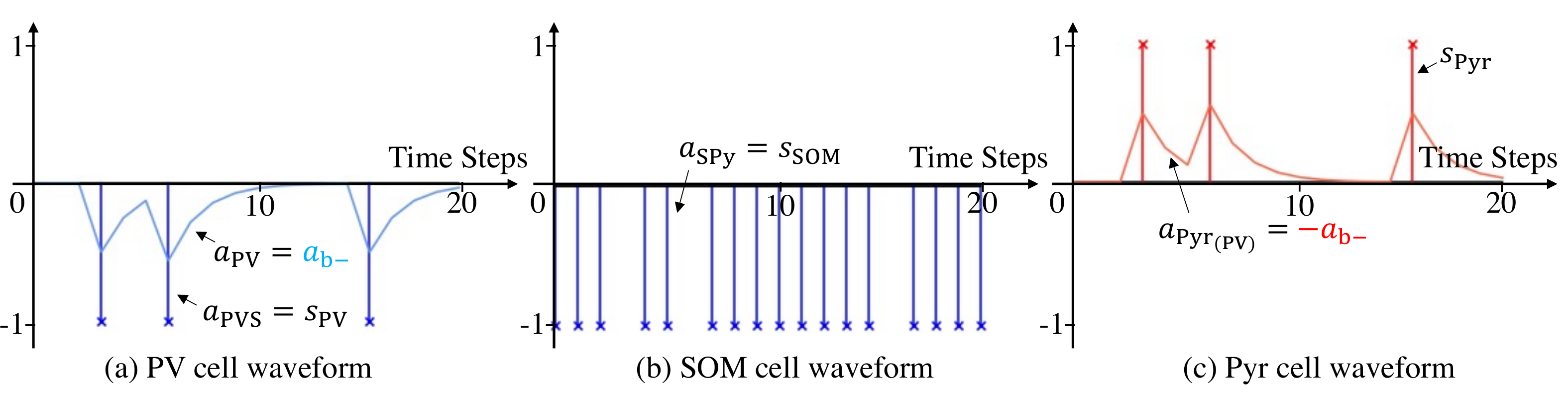}
    \caption{Inhibitory backward microcircuit 1 discrete time simulation. $\# {\rm Time~steps} = 20$, $\tau_s=2$}
    \label{fig:inh_back_micro_1_sim}
\end{figure}

The numerical simulation is similar as the excitatory backward microcircuit. All three cells also follow (\ref{eq:mem}) - (\ref{eq:a_time}). 
An example of all neurons' output waveforms is provided in Figure \ref{fig:inh_back_micro_1_sim}. In this simple example, the two fixed synapses have weights $w_{\rm PVS} = w_{\rm SPy} = 1$ and the same time constant $\tau_s = 1$. So according to (\ref{eq:a_time}), each spike emitted by both PV and SOM cells induces PSC $a_{\rm PVS} = a_{\rm SPy} = -1$ on that time-step. The SOM cell has $\tau_m = 1$, threshold $\vartheta = 0.5$, $I_{bias{\rm (SOM)}} = 1$. The range of these variables needs to satisfy: $(w_{\rm PVS}\cdot a_{\rm PVS} + I_{bias{\rm (SOM)}})<\vartheta < I_{bias{\rm (SOM)}}$. The Pyr cell has $\tau_m = 1$, threshold $\vartheta = 0.5$, $I_{bias{\rm (Pyr)}}=1$. The range of these variables needs to satisfy: $(w_{\rm SPy}\cdot a_{\rm SPy} + I_{bias{\rm (Pyr)}})<\vartheta < I_{bias{\rm (Pyr)}}$.

\subsubsection{Inhibitory backward microcircuit 2}
\begin{figure}[ht]
    \centering
    \includegraphics[width=0.99\textwidth]{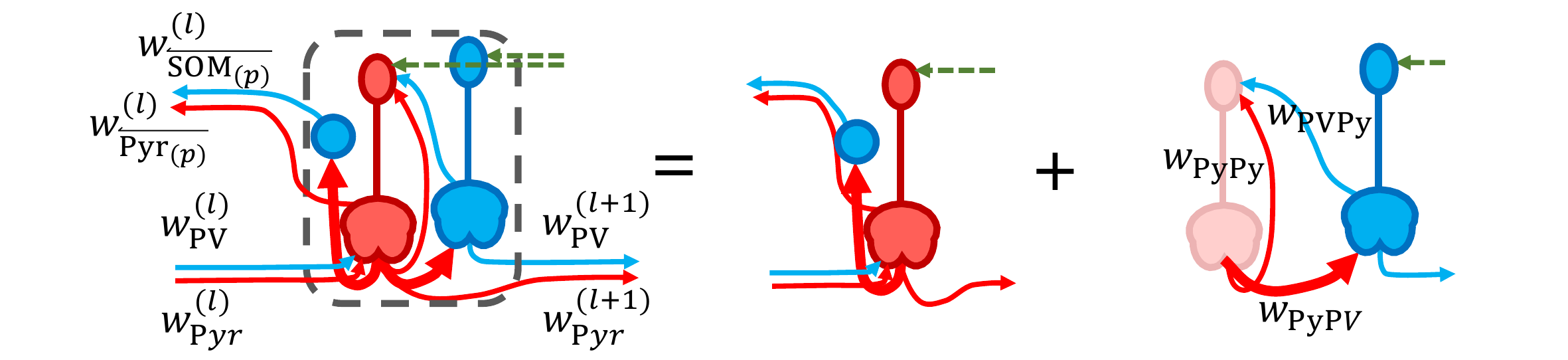}
    \caption{Inhibitory backward microcircuit 2}
    \label{fig:inh_back_micro_2}
\end{figure}

As shown in Figure \ref{fig:inh_back_micro_2}, we decompose the whole microcircuit into two parts. The first part is the excitatory backward microcircuit where a Pyr neuron backpropagates its error with the help of a pairing SOM neuron. The second part is newly added PV neuron and its connections.

A PV neuron only receives forward inputs from a paired Pyr neuron:
\begin{equation}
    I^{(l)}_{{\rm PV}_i}(t) = w^{(l)}_{{\rm PyPV}_i}\cdot a^{(l)}_{{\rm PyPV}_i}(t) 
\end{equation}
This fixed strong connection is designed to synchronize the firing behavior of a PV neuron with its paired Pyr neuron. Naming the forward PSCs of the $i$th Pyr and PV neurons locate in layer $(l)$ as $a^{(l)}_{{\rm Pyr}_i}$ and $a^{(l)}_{{\rm PV}_i}$, we have 
\begin{equation}
    a^{(l)}_{{\rm PV}_i} = - a^{(l)}_{{\rm Pyr}_i}.
    \label{eq:inh_mic2_pyr_pv}
\end{equation}
The other two connections propagates the PV cell's error signals (via $w_{\rm PVPy}$), and cancel out self-generated inhibitory input (via autapses $w_{\rm PyPy}$) to the Pyr cell's apical dendrites. The total inputs of the Pyr neuron's apical dendrites is the top-down input $I^{(l)}_{a_i}$ it originally received without the PV neuron, and the two new inputs from the PV neuron:
\begin{equation}
\begin{aligned}
    I^{(l)}_{a{\rm (Pyr)}_i} &= I^{(l)}_{a_i} + \left(w^{(l)}_{{\rm PyPy}_i}\cdot a^{(l)}_{{\rm PyPy}_i}\right) + \left( w^{(l)}_{{\rm PVPy}_i}\cdot a^{(l)}_{{\rm PVPy}_i}\right)\\
    &= I^{(l)}_{a_i} + \left(w^{(l)}_{{\rm PyPy}_i}\cdot a^{(l)}_{{\rm Pyr}_i}\right) + \left( w^{(l)}_{{\rm PVPy}_i}\cdot a^{(l)}_{{\rm PVPy}_i}\right)
\end{aligned}
    \label{eq:inh_mic2_pyr_apical}
\end{equation}
The anti-Hebbian rule is able to drive $w_{\rm PVPy} \approx w_{\rm PyPy}$ as shown in Section \ref{sec:anti_exp}. We continue the analysis assuming these two weights are converged to $w_{\rm au}$ (
${\rm au}$ stands for autapses). 

According to (\ref{eq:error}), we have:
\begin{equation}
    a^{(l)}_{{\rm PVPy}_i} = a^{(l)}_{{\rm PV}_i}-\sigma'(u) I^{(l)}_{a{\rm (PV)}_i}
    \label{eq:inh_mic2_pvpy}
\end{equation}
where $I^{(l)}_{a{\rm (PV)}_i}$ is the total apical input currents of the PV neuron. Bring (\ref{eq:error}), (\ref{eq:inh_mic2_pyr_pv}) and (\ref{eq:inh_mic2_pvpy}) into (\ref{eq:inh_mic2_pyr_apical}), we have:
\begin{equation}
\begin{aligned}
        I^{(l)}_{a{\rm (Pyr)}_i} &= I^{(l)}_{a_i} + \left(w^{(l)}_{{\rm PyPy}_i}\cdot a^{(l)}_{{\rm Pyr}_i}\right) + \left( w^{(l)}_{{\rm PVPy}_i}\cdot a^{(l)}_{{\rm PVPy}_i}\right)\\
        &= I^{(l)}_{a_i} + \left(w^{(l)}_{{\rm au}_i}\cdot a^{(l)}_{{\rm Pyr}_i}\right) + \left[ w^{(l)}_{{\rm au}_i}\cdot (a^{(l)}_{{\rm PV}_i}-\sigma'(u^{(l)}_{{\rm PV}_i}) I^{(l)}_{a{\rm (PV)}_i})\right]\\
        &= I^{(l)}_{a_i} - w^{(l)}_{{\rm au}_i}\sigma'(u^{(l)}_{{\rm PV}_i}) I^{(l)}_{a{\rm (PV)}_i}
\end{aligned}
\end{equation}
The Pyr backward signal $a_{\overleftarrow{+}_i}^{(l)}$ is:
\begin{equation}
\begin{aligned}
    a_{\overleftarrow{+}_i}^{(l)} &= a^{(l)}_{{\rm Pyr}_i} + \sigma'(u^{(l)}_{{\rm Pyr}_i}) I^{(l)}_{a{\rm (Pyr)}_i}\\
    &= a^{(l)}_{{\rm Pyr}_i} + \sigma'(u^{(l)}_{{\rm Pyr}_i}) \left(I^{(l)}_{a_i} - w^{(l)}_{{\rm au}_i}\sigma'(u^{(l)}_{{\rm PV}_i}) I^{(l)}_{a{\rm (PV)}_i}\right)\\
    &\approx a^{(l)}_{{\rm Pyr}_i} + \sigma'(u^{(l)}_{{\rm Pyr}_i}) \left(-\frac{\partial L}{\partial a^{(l)}_{{\rm Pyr}_i}} + w^{(l)}_{{\rm au}_i}\sigma'(u^{(l)}_{{\rm PV}_i})\frac{\partial L}{\partial a^{(l)}_{{\rm Pyr}_i}}\right) \\
    &= a^{(l)}_{{\rm Pyr}_i} - \sigma'(u^{(l)}_{{\rm Pyr}_i})\frac{\partial L}{\partial a^{(l)}_{{\rm Pyr}_i}}\left(1 - w^{(l)}_{{\rm au}_i}\sigma'(u^{(l)}_{{\rm PV}_i})\right)
    \label{eq:inh_mic2_pyr_back}
\end{aligned}
\end{equation}

The approximation comes from the setting of SFA, where the apical dendrites current approximately stands for the negative derivative of loss w.r.t neurons' output PSCs. The term $ w^{(l)}_{{\rm au}_i}$ is a constant, which can be absorbed into the random backward weights so it does not affect the error propagation. From (\ref{eq:inh_mic2_pyr_back}), we know that the propagation of errors in a Pyr neuron $\partial L/\partial u^{(l)}_{{\rm Pyr}_i} = \sigma'(u^{(l)}_{{\rm Pyr}_i})\frac{\partial L}{\partial a^{(l)}_{{\rm Pyr}_i}}$ works exactly as the excitatory backward microcircuit. Yet the errors of a PV neuron goes through the membrane dependent surrogate functions twice, which differs from the typical backpropagation dataflow. Because all forward inputs from previous layers are connect to the Pyr neuron, the only membrane dependent surrogate function needed is Pyr neuron's $\sigma'(u^{(l)}_{{\rm Pyr}_i})$. 

\begin{figure}[ht]
    \centering
    \includegraphics[width=0.99\textwidth]{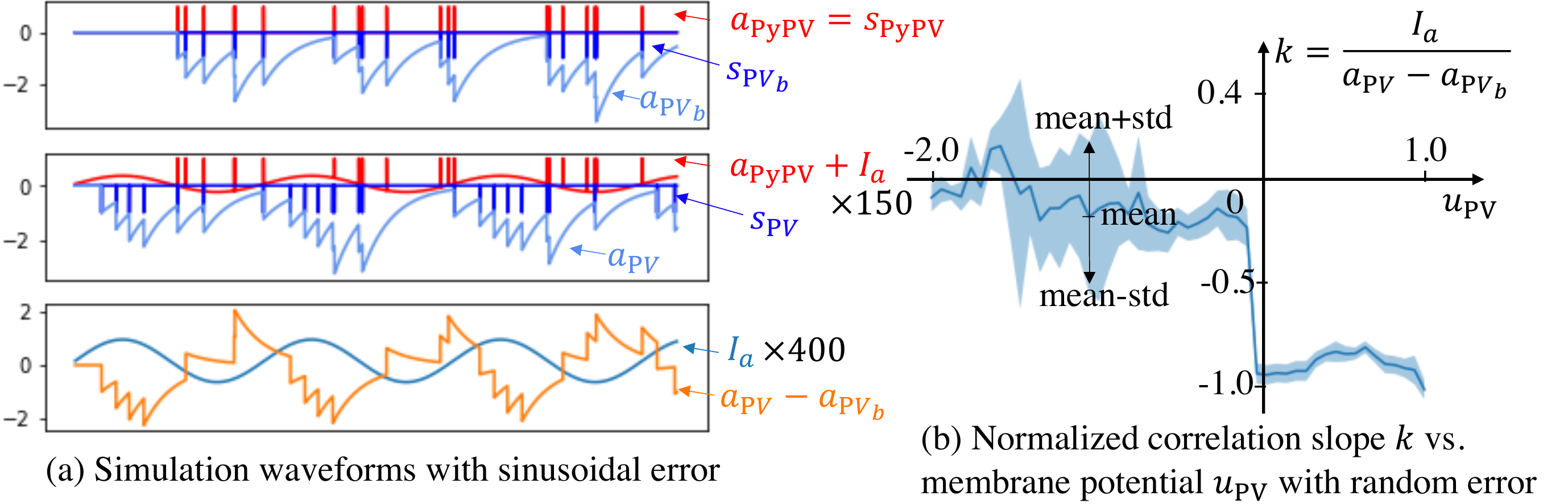}
    \caption{Inhibitory backward microcircuit 2 simulation}
    \label{fig:inh_back_micro_2_sim}
\end{figure}

We solve this issue by setting the time constant of PV neurons' membrane potential $\tau_m$ as a large number, or simply use the integrate-and-fire (IF) model for all PV neurons. Under this setting, the additional term $\sigma'(u^{(l)}_{{\rm PV}_i}$ is a constant. As shown in Figure \ref{fig:inh_back_micro_2_sim}, we add an experimental simulation to demonstrate the dynamics of this microcircuit.


In Figure \ref{fig:inh_back_micro_2_sim} (a)'s first subplot, we show that the firing behavior of PV and Pyr neurons synchronized successfully. As previous microcircuit, $\tau_s=1$ for the fixed weight synapse $w_{\rm PyPV}=1$. We set PV's threshold $\vartheta=0.9$ to make sure that each Pyr spike can charge the PV's membrane potential to move over its threshold. In the second subplot, we add a small sinusoidal signal simulating apical dendrites injected error of the PV neuron. For visualization purpose, we multiplied the summed signal by a factor of 150 when plotting. One can observe that the firing behavior of the PV neuron changes. In the third subplot, we plot the sinusoidal signal together with the change of the PV's output PSC. The two signals are clearly negatively correlated (for visualization purpose, we multiplied the error signal by a factor of 400). In Figure \ref{fig:inh_back_micro_2_sim} (b), we repeat this process 100 times, and plot the correlation slope w.r.t different value of membrane potential at each time step. The resulting curve is almost flat when $u_{\rm PV}>0$, so it satisfied our assumption that $\sigma'(u^{(l)}_{{\rm PV}_i}$ is a constant.

\subsection{Inhibitory backward microcircuit 3}
\begin{figure}[ht]
    \centering
    \includegraphics[width=0.99\textwidth]{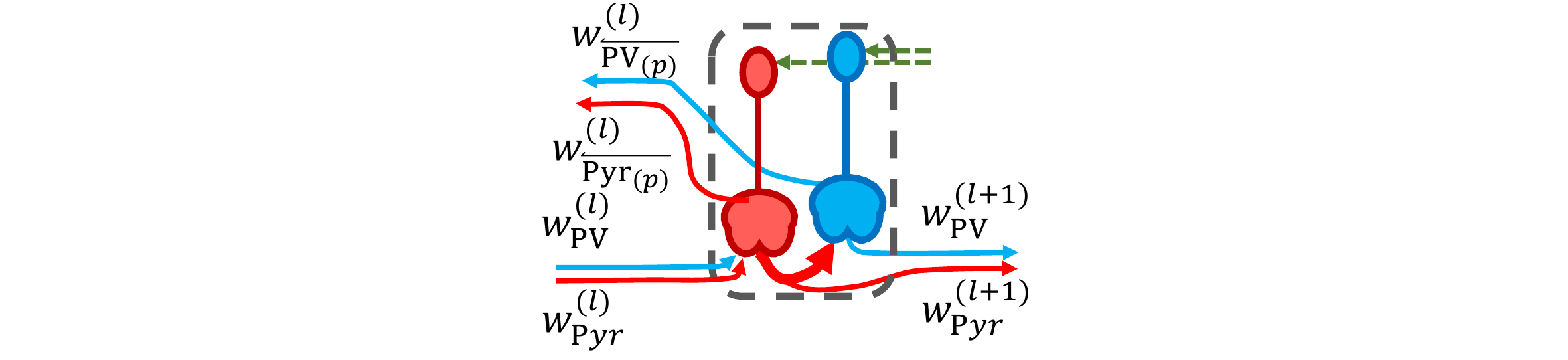}
    \caption{Inhibitory backward microcircuit 3}
    \label{fig:inh_back_micro_3}
\end{figure}

The third microcircuit architecture is shown in Figure \ref{fig:inh_back_micro_3}. By connecting each PV neuron from a Pyr neuron through a strong fixed connection as in inhibitory backward microcircuit 2, The two neurons are synchronized so they can canceling each other's self-generated PSCs themselves 
Like the inhibitory backward microcircuit 2, a PV neuron only receives forward inputs from a paired Pyr neuron:
\begin{equation}
    I^{(l)}_{{\rm PV}_i}(t) = w^{(l)}_{{\rm PyPV}_i}\cdot a^{(l)}_{{\rm PyPV}_i}(t) 
\end{equation}
Naming the partial derivative of loss w.r.t the membrane potential of their next layer as $\partial L/\partial u^{(l+1)}$, the apical dendrites input currents of Pyr and PV neurons are $\boldsymbol{B}^{(l+1)}_{\rm Pyr}\cdot (\partial L/\partial u^{(l+1)})$ and $\boldsymbol{B}^{(l+1)}_{\rm PV}\cdot (\partial L/\partial u^{(l+1)})$. They only differ on the magnitude of the random values in the backward matrix $\boldsymbol{B}$. Therefore, the information they carried is duplicated. Using the one located in the Pyr neuron's apical dendrites only to adjust the input synaptic weights is sufficient.

\section{Equivalence to SFA}

\section{Detailed experimental setups}
When running on the MNIST and CIFAR10 datasets, we use inhibitory backward microcircuit 2 or 3. Both of which requires a pairing behavior between Pyr and PV neurons within the same layer. The inhibitory backward microcircuit 1 performs poorer than the other two, which gained 98.35\% best accuracy on the MNIST dataset, and failed to gain any meaningful results on the CIFAR10 dataset (accuracy always near 10\%).

For a fair comparison, we divided each layer half-half: half of the neurons are excitatory Pyr neurons, and the other half are inhibitory PV neurons. In which case the neurons number are the same.  We violate the well-observed 80\%/20\% Exc/Inh ratio. We hypothesize that neurons involved in the cortical structure not only support computation, but also facilitate many other important functions, such as maintaining stability. However, our work only focuses on computational/learning functions. 

Moreover, if we change the error backpropagation from layer-by-layer (FA) to direct feedback (DFA), the ratio can be brought closer to 80\%/20\%. In DFA, only neurons in the output layer broadcast errors backwards, so only each neuron in the output layer needs one-to-one pairing of an auxiliary neuron. FA and DFA are just two error propagation schemes. They can all be build on our bio-plausible fabrics.

Ideally, convolutional layer and pooling layer are not bio-plausible. So we also tested the network performance with only fully-connected layers. With no hidden layer, the network gains 62.3\% accuracy; with one hidden layer, which has 100 Pyr neurons and 100 PV neurons, the network gains 90.4\% accuracy. With two hidden layers, each has 100 Pyr and 100 PV neurons, the network gains 94.8\% accuracy.

Since both datasets' inputs are normalized, which means there are data in both positive and negative. However, negative values cannot charge a spiking neuron to generate a spike. Therefore, we allow the weights of the input layer to be both positive and negative.

\subsection{Training setups for different datasets} 
\subsubsection{MNIST}
The MNIST dataset \cite{lecun1998mnist} contains 60,000 training images and 10,000 testing images.  We set the batch size to 64, the number of training epochs to 200, and the learning rate to 0.0005 for the adopted AdamW optimizer \cite{loshchilov2017decoupled}. The images were converted to  continuous-valued multi-channel  currents applied as the inputs to the SNN under training.    Moreover, data augmentations using  RandomCrop and RandomRotation were applied to improve performance \cite{shorten2019survey}. 
\subsubsection{CIFAR10}
The CIFAR10 dataset \cite{krizhevsky2009learning} contains 50,000 training images and 10,000 test images.
We trained the SNN for 600 epochs with a batch size of 50 and a learning rate 0.0005 for the AdamW optimizer \cite{loshchilov2017decoupled}. 
The input image coding strategy used for the MNIST dataset was also adopted here. Moreover, data augmentations including RandomCrop, ColorJitter, and RandomHorizontalFlip \cite{shorten2019survey} were applied. The convolutional layers were initialized using the kaiming uniform initializer \cite{he2015delving}, and the linear layers were initialized using the kaiming normal initializer \cite{he2015delving}.

\subsection{Assembly competition experiment}

When running the experiment, we set the time constant of both membrane potential and post synaptic current as 1 time step. The learning rate = 0.1, and the update rule is
\begin{equation}
     \Delta \boldsymbol{W} = \eta * (\boldsymbol{a}_{\rm Pyr} \otimes \boldsymbol{a}_{\rm SOM} - 0.5) \odot \boldsymbol{W} \odot (1-\boldsymbol{W})
\end{equation}
In the above equation, $\otimes$ is the outer product, $\odot$ is the elementwise product. Each element in $\boldsymbol{a}_{\rm Pyr}$ equal to $1$ when the Pyr neuron fires, and equal to $0$ when not fire.  Each element in $\boldsymbol{a}_{\rm SOM}$ equal to $-1$ when the Pyr neuron fires, and equal to $0$ when not fire.
$\boldsymbol{W}$ is initialized randomly following a uniform distribution $U(0, w_{\rm max})$ ranging from zero to the max allowed weights $w_{\rm max}$. The value of $w_{\rm max}$ is free to choose, and would not impact the success of simulation in certain range (In our setting, $w_{\rm max}$ can ranging in $[\vartheta,3\vartheta]$, where $\vartheta=1$ is the SOM cell's threshold).   Name a weight from Pyr cell j to SOM cell i as $w_{ij}$, we have $\Delta w_{ij} = a^+ w_{ij}(w_{\rm max}-w_{ij})$, if $i$ and $j$ fires together, and $\Delta w_{ij} = a^- w_{ij}(w_{\rm max}-w_{ij})$ otherwise (We set $a^+=0.05$, and $a^-=-0.05$. These two constants does not affect the simulation much, and even imbalance values work). All weights are soft-bounded by the dynamic learning rate $w_{ij}(w_{\rm max}-w_{ij})$ \cite{kheradpisheh2018stdp}, which has smaller value when a weight is moving closer to the boundary.




\section{Limitations of this work}
\label{sec:app_lim}
\subsection{Violation of the well-observed Excitatory/Inhibitory neurons' 80\%/20\% ratio}
The main idea of our current framework is based on layer-by-layer backpropagation. We notify that our framework violates the 80\%/20\% ratio. We hypothesize that neurons involved in the cortical structure not only support computation, but also facilitate many other important functions, such as maintaining stability. However, our work only focuses on computational/learning functions. 

Moreover, if we change the error backpropagation from layer-by-layer (FA) to direct feedback (DFA), the ratio can be brought closer to 80\%/20\%. In DFA, only neurons in the output layer broadcast errors backwards, so only each neuron in the output layer needs one-to-one pairing of an auxiliary neuron. FA and DFA are just two error propagation schemes. They can all be build on our bio-plausible fabrics.

\subsection{Neurons' anatomies are not fully justified}
The ability for the various dendritic regions to engage in distinct computations from their soma is not fully accepted and is the subject of much ongoing research \cite{beaulieu2019widespread, francioni2019high}. We do not claim our model including two-compartments Pyr \& PV cells and single compartment SOM cells is \textbf{the} computational framework of the real brain, but only a possible simplified modeling choice.

\end{document}